\documentclass[
a4paper,
man,
british
]{apa7}

\usepackage{changes}
\usepackage{cancel}
\usepackage[british]{babel}
\usepackage[utf8]{inputenc}
\usepackage{epstopdf}
\usepackage{csquotes}
\usepackage[flushleft]{threeparttable}
\usepackage{multirow}
\usepackage[
style = apa,
backend = biber,
sortcites = true,
sorting = nyt,
hyperref = true,
backref = false,
]{biblatex}
\hypersetup{
colorlinks = true,
linkcolor = black,
anchorcolor = black,
citecolor = black,
filecolor = black,
urlcolor = blue
}
\usepackage{float}
\usepackage{placeins}
\usepackage{xcolor}
\usepackage[toc, page]{appendix}
\usepackage{lscape}
\usepackage{afterpage}
\usepackage{esvect}
\usepackage{amsmath}
\usepackage{ragged2e}
\justifying\let\raggedright\justifying
\usepackage{enumitem}
\usepackage{makecell}
\DeclareLanguageMapping{british}{british-apa}
\usepackage[nodisplayskipstretch]{setspace}
\usepackage{subcaption} 
\usepackage{rotating}
\usepackage{geometry}
\let \citeA \textcite
\let \cite \parencite

\newcommand{\figurehere}[1]{\begin{center}%
\vspace{-2mm}
=========================\\%
Insert Figure #1 about here\\%
=========================\\%
\vspace{-2mm}
\end{center}}
\newcommand{\tablehere}[1]{\begin{center}%
\vspace{-2mm}
=========================\\%
Insert Table #1 about here\\%
=========================\\%
\vspace{-2mm}
\end{center}}

\usepackage{array}
\newcommand{\PreserveBackslash}[1]{\let\temp=\\#1\let\\=\temp}
\newcolumntype{C}[1]{>{\PreserveBackslash\centering}p{#1}}
\newcolumntype{R}[1]{>{\PreserveBackslash\raggedleft}p{#1}}
\newcolumntype{L}[1]{>{\PreserveBackslash\raggedright}p{#1}}

\title{Tutorial on Using Machine Learning and Deep Learning Models for Mental Illness Detection}
\shorttitle{Data-driven Methods to Identify Mental Illness}

\authorsnames[1,2,2,3,4,5,1,5]{%
  Yeyubei Zhang, Zhongyan Wang, Zhanyi Ding, Yexin Tian, 
  Jianglai Dai, Xiaorui Shen, Yunchong Liu, Yuchen Cao
}
\authorsaffiliations{%
  {University of Pennsylvania, School of Engineering and Applied Science}, 
  {New York University, Center for Data Science}, 
  {Georgia Institute of Technology, College of Computing}, 
  {University of California, Berkeley, Department of EECS}, 
  {Northeastern University, Khoury College of Computer Science}
}

\abstract{Social media has become an important source for understanding mental health, providing researchers a way to detect conditions like depression from user-generated posts. This tutorial provides practical guidance to address common challenges in applying machine learning and deep learning methods for mental health detection on these platforms. It focuses on strategies for working with diverse datasets, improving text preprocessing, and addressing issues such as imbalanced data and model evaluation. Real-world examples and step-by-step instructions demonstrate how to apply these techniques effectively, with an emphasis on transparency, reproducibility, and ethical considerations. By sharing these approaches, this tutorial aims to help researchers build more reliable and widely applicable models for mental health research, contributing to better tools for early detection and intervention.}


\authornote{
Correspondence concerning this article should be addressed to Yuchen Cao, Northeastern University, E-mail: cao.yuch@northeastern.edu}

\begin{document}
\maketitle

\section{Introduction}

Mental health disorders, especially depression, have become a significant concern worldwide, affecting millions of individuals across diverse populations \cite{WHO2020}. Early detection of depression is crucial, as it can lead to timely treatment and better long-term outcomes. In today’s digital age, social media platforms such as X(Twitter), Facebook, and Reddit provide a unique opportunity to study mental health. People often share their thoughts and emotions on these platforms, making them a valuable source for understanding mental health patterns \cite{Choudhury2013, Guntuku2017}.

Recent advances in computational methods, particularly machine learning (ML) and deep learning (DL), have shown promise in analyzing social media data to detect signs of depression. These techniques can uncover patterns in language use, emotions, and behaviors that may indicate mental health challenges \cite{Shatte2020, Yazdavar2020}. 

However, applying these methods effectively is not without challenges. A recent systematic review \cite{cao2024mental} highlighted issues such as a lack of diverse datasets, inconsistent data preparation, and inadequate evaluation metrics for imbalanced data \cite{Hargittai2015, Helmy2024}—problems that have also led to inaccuracies in other domains (e.g., \cite{gao2024survey, weng2024ai}). Similarly, Liu et al. \cite{liu2024systematic} identified additional linguistic challenges in ML approaches for detecting deceptive activities on social networks, including biases from insufficient linguistic preprocessing and inconsistent hyperparameter tuning, all of which are pertinent to mental health detection. Moreover, complementary insights from related fields underscore the need for continuous improvements in robust model development \cite{bi2024decoding,zhao2024minimax,tao2023meta,xu2025robust}.

This tutorial is designed to address these gaps by guiding readers through the steps necessary to create reliable and accurate models for depression detection using social media data. It focuses on practical techniques to:
\begin{itemize}
\item Collect and preprocess data, including handling language challenges like sarcasm or negations.
\item Build and optimize models with attention to tuning and evaluation.
\item Use appropriate metrics for datasets where depressive posts are a minority.
\end{itemize}

Our goal is to provide a clear, step-by-step approach that researchers and practitioners can use to improve their methods. By addressing common challenges in this field, we hope to encourage more robust and ethical use of technology for improving mental health outcomes.

\section{Method}
This section provides a comprehensive overview of the methodological framework employed in this study, detailing the processes for data preparation, model development, and evaluation metrics. All analyses and model implementations were conducted using Python 3, leveraging popular libraries such as \texttt{pandas} for data manipulation, \texttt{scikit-learn} for machine learning, \texttt{PyTorch} for deep learning, and \texttt{Transformers} for pre-trained language models. These tools enabled efficient preprocessing, hyperparameter optimization, and performance evaluation. The following subsections elaborate on the key steps and methodologies involved in the study.

\subsection{Data Preparation}
\subsubsection{Data Sources and Collection Methods}
A sufficiently representative dataset is essential for machine-learning-based mental health detection. This study utilized the Sentiment Analysis for Mental Health dataset, available on \href{https://www.kaggle.com/datasets/suchintikasarkar/sentiment-analysis-for-mental-health/data}{Kaggle}. The dataset integrates textual content from multiple repositories focused on mental health topics, including depression, anxiety, stress, bipolar disorder, personality disorders, and suicidal ideation. The primary sources of these data are social media platforms such as Reddit, Twitter, and Facebook, where individuals frequently discuss personal experiences, emotional states, and mental health concerns.

The dataset was originally compiled using platform-specific APIs (e.g., Reddit, Twitter, and Facebook) and web scraping tools, allowing for the collection of substantial volumes of publicly available text data. After the acquisition, duplicates were removed, irrelevant and spam content was filtered, and mental health labels were standardized to ensure consistency across repositories. Personal identifiers were removed to safeguard privacy and ensure compliance with ethical guidelines for data usage. The final dataset was consolidated into a structured CSV file with unique identifiers for each entry.

Although the dataset combines data from multiple platforms to provide a diverse corpus, it is not free from limitations. Differences in platform demographics, such as age, cultural background, and communication styles, may affect the generalizability of models trained on this data. Additionally, linguistic variability, including colloquialisms, slang, and code-switching, reflects the informal nature of social media communication. While this diversity enriches the dataset, it also presents challenges for natural language processing (NLP) techniques, particularly in tokenization and embedding generation. To address these complexities, the preprocessing pipeline was designed to handle diverse linguistic patterns and balance class distributions where needed.

\subsubsection{Data Preprocessing}
A standardized preprocessing pipeline was applied to prepare the dataset for training both machine learning (ML) and deep learning (DL) models. These steps ensured consistency in data handling while accommodating the unique requirements of each modeling approach:
\vspace{-3mm}
\begin{itemize}
\item \textbf{Text Cleaning:} Social media text often contains noise such as URLs, HTML tags, mentions, hashtags, special characters, and extra whitespace. These elements were systematically removed using regular expressions to create cleaner input for both ML and DL models.
\vspace{-1.5mm}
\item \textbf{Lowercasing:} All text was converted to lowercase to maintain uniformity across the dataset and minimize redundancy in text representation.
\vspace{-1.5mm}
\item \textbf{Stopword Removal:} Commonly used words that provide little semantic value (e.g., “the,” “and,” “is”) were excluded using the stopword list available in the Natural Language Toolkit (NLTK) \cite{nltk_toolkit}, reducing noise while retaining meaningful content.
\vspace{-1.5mm}
\item \textbf{Lemmatization:} Words were reduced to their base forms (e.g., “running,” “ran,” “runs” → “run”) using NLTK's Lemmatizer. This step normalized variations of words, aiding both feature extraction and embedding generation.
\end{itemize}
\vspace{-3mm}

The dataset was divided into training, validation, and testing subsets using a two-step random sampling process with a fixed random seed to ensure reproducibility. First, 20\% of the data was set aside as the test set. The remaining 80\% was then further divided into a training set (60\% of the original data) and a validation set (20\% of the original data). This split ensured that the models were trained on the majority of the data while reserving separate subsets for hyperparameter tuning and final performance evaluation.

\subsubsection{Class Labeling}
The dataset’s class labels were prepared as follows: (1) For \textbf{multi-class classification}, the labels included six categories: Normal, Depression, Suicidal, Anxiety, Stress, and Personality Disorder. (2) For \textbf{binary classification}, the labels were grouped into two classes: Normal and Abnormal.

\subsubsection{Feature Transformation for ML Models}
For ML models, an additional step, TF-IDF Vectorization, was necessary to transform the text into structured features. The cleaned text was converted into numerical representations using Term Frequency–Inverse Document Frequency (TF-IDF), which captured term frequencies while down-weighting overly frequent words. The vectorizer was configured to extract up to 1,000 features and account for both unigrams and bigrams (n-gram range: 1–2).

\subsubsection{Code Availability}
The code for data preprocessing, including text cleaning, class labeling, and dataset splitting, is publicly available on GitHub (the link will be provided upon acceptance).

\subsection{Model Development}
This study employed a range of machine learning (ML) and deep learning (DL) models to analyze and classify mental health statuses based on textual data. Each model was selected to explore specific aspects of the data, from linear interpretability to handling complex patterns and long-range dependencies. Detailed implementation code for all models, including hyperparameter tuning and evaluation, is available on GitHub. Below, we provide an overview of each model, its methodology, and its performance in the context of binary and multi-class mental health classification tasks.

\subsubsection{Logistic Regression}
Logistic regression is one of the most widely used methods for classification tasks and has long been employed in social science and biomedical research \cite{hosmer2000applied, ding2025efficientpowerfultradeoffsmachine}. In the context of mental health detection, it provides a straightforward yet interpretable modeling framework, translating linear combinations of predictors (e.g., term frequencies, sentiment scores, and linguistic features) into estimated probabilities of class membership through the logit function.

The logistic regression model predicts the probability of a binary outcome using the following expression:
\begin{equation}\nonumber
\hat{y} = \frac{1}{1 + \exp(-\mathbf{w}^\top \mathbf{x} - b)},
\end{equation}
where $\hat{y}$ represents the predicted probability, $\mathbf{w}$ is the vector of model coefficients, $\mathbf{x}$ is the feature vector, and $b$ is the bias term. For multi-class classification, the model generalizes to predict probabilities for $K$ classes using the softmax function:
\begin{equation}\nonumber
P(y = k \mid \mathbf{x}) = \frac{\exp(\mathbf{w}_k^\top \mathbf{x} + b_k)}{\sum_{j=1}^{K} \exp(\mathbf{w}_j^\top \mathbf{x} + b_j)},
\end{equation}
where $k \in \{1, \dots, K\}$ represents the class index.

Both binary and multi-class logistic regression models were optimized using cross-entropy loss during training and configured to converge with a maximum iteration limit of 1,000. Regularization was applied to prevent overfitting, using $\ell_2$ (ridge) regularization, which penalizes large coefficients by adding their squared magnitude to the loss function:
\begin{equation}\nonumber
\mathcal{L} = - \frac{1}{n} \sum_{i=1}^{n} \left[ y_i \log(\hat{y}_i) + (1 - y_i) \log(1 - \hat{y}_i) \right] + \lambda \|\boldsymbol{\beta}\|_2^2,
\end{equation}
where $\lambda$ controls the regularization strength, $y_i$ is the true label, $\hat{y}_i$ is the predicted probability, and $\boldsymbol{\beta}$ represents the model coefficients.

Hyperparameter tuning was conducted using a grid search across several parameters. The regularization strength (\(C\)), which is the inverse of the regularization parameter \(\lambda\), was tested with values such as 0.1, 1, and 10. Various optimizers, including \texttt{liblinear} (Library for Large Linear Classification), \texttt{lbfgs} (Limited-memory Broyden–Fletcher–Goldfarb–Shanno), and \texttt{saga} (Stochastic Average Gradient Augmented), were evaluated for optimization. To address class imbalance, the \texttt{class\_weight} parameter was explored with options for \texttt{balanced} and \texttt{None}. For multi-class tasks, the \texttt{multinomial} strategy was employed, while the \texttt{one-vs-rest} strategy was implicitly applied for binary classification scenarios.

For both binary and multi-class tasks, the weighted F1 score was used as the primary evaluation metric, ensuring balanced performance across categories, including minority classes. A combined grid search configuration was applied for both tasks, as their hyperparameter spaces largely overlapped. The best configurations effectively handled class imbalance using the \texttt{class\_weight=`balanced'} parameter, yielding robust performance on the validation and test sets.

The logistic regressions were implemented using the \texttt{LogisticRegression} class from the \texttt{scikit-learn} library. Detailed implementation code for logistic regression, including hyperparameter tuning and evaluation, is available on GitHub.

\subsubsection{Support Vector Machine (SVM)}
Support Vector Machines (SVMs) are supervised learning models that are widely used for both classification and regression tasks. Originally introduced by \citeA{cortes1995support}, SVMs aim to find the optimal hyperplane that maximizes the margin between data points of different classes. The margin is defined as the distance between the closest data points (support vectors) from each class to the hyperplane. By maximizing this margin, SVMs achieve better generalization for unseen data.

For a linearly separable dataset, the decision boundary is defined as:
\begin{equation}\nonumber
f(\mathbf{x}) = \mathbf{w}^T \mathbf{x} + b,
\end{equation}
where $\mathbf{w}$ is the weight vector, $\mathbf{x}$ is the feature vector, and $b$ is the bias term. The optimal hyperplane is determined by solving the following optimization problem:
\begin{align}
\min_{\mathbf{w}, b} & \quad \frac{1}{2} \|\mathbf{w}\|^2, \nonumber\\
\text{subject to} & \quad y_i (\mathbf{w}^T \mathbf{x}_i + b) \geq 1, \quad i = 1, \dots, N,\nonumber
\end{align}
where $y_i \in \{-1, +1\}$ are the class labels.

For datasets that are not linearly separable, the optimization problem is modified to include a penalty for misclassifications:
\begin{align}
\min_{\mathbf{w}, b, \xi} & \quad \frac{1}{2} \|\mathbf{w}\|^2 + C \sum_{i=1}^N \xi_i, \nonumber\\
\text{subject to} & \quad y_i (\mathbf{w}^T \mathbf{x}_i + b) \geq 1 - \xi_i, \quad \xi_i \geq 0, \quad i = 1, \dots, N,\nonumber
\end{align}
where $\xi_i$ are slack variables that allow for misclassifications, and $C > 0$ is the regularization parameter that controls the trade-off between maximizing the margin and minimizing classification errors.

Kernel methods enable SVMs to handle nonlinearly separable data by mapping the input features into a higher-dimensional space where linear separation becomes possible. This mapping is performed implicitly using a kernel function \( K(\mathbf{x}_i, \mathbf{x}_j) \), which computes the inner product in the transformed space:
\begin{equation}\nonumber
K(\mathbf{x}_i, \mathbf{x}_j) = \phi(\mathbf{x}_i)^T \phi(\mathbf{x}_j),
\end{equation}
where \( \phi(\cdot) \) represents the mapping function.

Several commonly used kernel functions are available, each suited for different data characteristics:

\vspace{-3mm}
\begin{enumerate}
\item \textbf{Linear Kernel:}
\begin{equation}\nonumber
K(\mathbf{x}_i, \mathbf{x}_j) = \mathbf{x}_i^T \mathbf{x}_j
\end{equation}
This kernel computes the dot product of the input vectors and is suitable for linearly separable data.
\vspace{-1.5mm}
\item \textbf{Polynomial Kernel:}
\begin{equation}\nonumber
K(\mathbf{x}_i, \mathbf{x}_j) = (\mathbf{x}_i^T \mathbf{x}_j + c)^d,
\end{equation}
where \( c \) is a constant and \( d \) is the degree of the polynomial. This kernel is useful for capturing polynomial feature interactions.
\vspace{-1.5mm}
\item \textbf{Radial Basis Function (RBF) Kernel:}
\begin{equation}\nonumber
K(\mathbf{x}_i, \mathbf{x}_j) = \exp\left(-\gamma \|\mathbf{x}_i - \mathbf{x}_j\|^2\right),
\end{equation}
where \( \gamma \) controls the influence of individual training samples. The RBF kernel is widely used for its flexibility in modeling complex, nonlinear patterns.
\vspace{-1.5mm}
\item \textbf{Sigmoid Kernel:}
\begin{equation}\nonumber
K(\mathbf{x}_i, \mathbf{x}_j) = \tanh(\alpha \mathbf{x}_i^T \mathbf{x}_j + c),
\end{equation}
where \( \alpha \) is a scaling factor and \( c \) is a bias term. This kernel is inspired by neural network activation functions and is suitable for data with specific characteristics.
\vspace{-1.5mm}
\item \textbf{Custom Kernels:}
Custom-defined kernels can be tailored for domain-specific tasks, offering flexibility for unique datasets or similarity metrics.
\end{enumerate}
\vspace{-3mm}

In this project, kernel selection was based on preliminary experiments, with the linear and radial basis function (RBF) kernels being the primary choice due to its ability to model complex, nonlinear relationships effectively.

For both binary and multi-class classification tasks, the same hyperparameter tuning strategy was employed. A grid search was conducted over the following hyperparameters:
\vspace{-3mm}
\begin{itemize}
\item Regularization parameter $C$: values of \{0.1, 1, 10\}.
\vspace{-1.5mm}
\item Kernel type: linear and RBF.
\vspace{-1.5mm}
\item Class weight: balanced or none.
\vspace{-1.5mm}
\item Gamma (for RBF kernel): scale and auto.
\end{itemize}
\vspace{-3mm}
The grid search aimed to identify the optimal combination of hyperparameters using the weighted F1 score as the primary evaluation metric. For multi-class classification, the one-vs-one strategy inherent to the \texttt{SVC} implementation was used.

The loss function for SVM is analogous to logistic regression, as both models minimize the cross-entropy loss during optimization. However, for SVM, the hinge loss is typically used for linear separable cases, defined as:
\begin{equation}\nonumber
\mathcal{L}_{\text{hinge}} = \frac{1}{N} \sum_{i=1}^N \max(0, 1 - y_i f(\mathbf{x}_i)).
\end{equation}

The SVM models were implemented with the \texttt{SVC} class from \texttt{scikit-learn}. Detailed implementation code for SVMs, including grid search and evaluation, is available on GitHub.

\subsubsection{Tree-Based Models}
Classification and Regression Trees (CARTs) are versatile tools used for analyzing categorical outcomes (classification tasks). The CART algorithm constructs a binary decision tree by recursively partitioning the data based on covariates, optimizing a predefined splitting criterion. For classification tasks, the quality of a split is typically evaluated using impurity measures such as Gini impurity or entropy \cite{bishop2006pattern}. The Gini impurity for a node is defined as:
\begin{equation}\nonumber
G = \sum_{i=1}^C p_i (1 - p_i),
\end{equation}
where \(p_i\) is the proportion of observations in class \(i\) at the given node, and \(C\) is the total number of classes.

Alternatively, entropy can be used to measure impurity:
\begin{equation}\nonumber
H = -\sum_{i=1}^C p_i \log(p_i),
\end{equation}
where \(p_i\) represents the same class proportions as in the Gini impurity formula. Lower impurity values indicate greater homogeneity within a node.

At each step, the algorithm selects the split that minimizes the weighted impurity of the child nodes. The impurity reduction for a given split is computed as:
\begin{equation}\nonumber
\Delta I = I_{\text{parent}} - \left( \frac{n_L}{n} I_L + \frac{n_R}{n} I_R \right),
\end{equation}
where \(I_{\text{parent}}\) is the impurity of the parent node, \(I_L\) and \(I_R\) are the impurities of the left and right child nodes, \(n_L\) and \(n_R\) are the number of observations in the left and right child nodes, and \(n\) is the total number of observations in the parent node.

The splitting process continues until one stopping criterion is met. Common criteria include: (1) a minimum number of samples in a node, (2) a maximum tree depth, and (3) No further reduction in impurity beyond a predefined threshold.

To address overfitting, pruning techniques \cite{breiman1984classification} are employed. Pruning reduces the tree size by removing splits that contribute minimally to predictive performance, improving the model's generalizability.

Due to their tendency to overfit, simple CART models were not evaluated in this project. Instead, ensemble methods like Random Forests and Gradient Boosted Trees, which combine multiple CART models, were used for improved robustness.

\paragraph{Random Forests}
Random Forests are ensemble learning methods that aggregate multiple decision trees parallelly to enhance classification performance. By building trees on bootstrap samples of the data and introducing random feature selection at each split, Random Forests reduce overfitting and improve generalization. Each tree is trained on a random bootstrap sample, where data points are sampled with replacement from the original dataset, meaning some observations may appear multiple times in the training sample, while others are excluded. Additionally, Random Forests introduce randomness during the tree-building process by selecting a random subset of covariates at each split instead of considering all available covariates. This randomization decorrelates the trees, reduces variance, and enhances the model's robustness. For classification tasks, the final prediction is determined by majority voting across all trees \cite{breiman2001random}.

To further mitigate overfitting, each tree in the Random Forest is grown to its full depth without pruning, fitting the bootstrap sample as accurately as possible. Hyperparameters such as the number of trees (\texttt{n\_estimators}), the maximum depth of each tree (\texttt{max\_depth}), and the minimum samples required to split a node (\texttt{min\_samples\_split}) or form a leaf (\texttt{min\_samples\_leaf}) play a critical role in balancing bias and variance. The parameter \texttt{class\_weight}, when set to \texttt{`balanced'}, adjusts weights inversely proportional to class frequencies, effectively addressing the class imbalance.

A grid search approach was employed to optimize key hyperparameters for both binary and multi-class classification tasks. The parameter grid explored values such as 50, 100, and 200 for the number of trees (\texttt{n\_estimators}), depths of 10, 20, or unrestricted (\texttt{None}) for \texttt{max\_depth}, and split criteria (\texttt{min\_samples\_split} and \texttt{min\_samples\_leaf}) to control tree complexity. The weighted F1 score served as the primary evaluation metric to account for imbalances in the dataset. For the binary classification task, the best-performing model, determined through validation, effectively handled class imbalance and demonstrated robust predictive performance for distinguishing between Normal and Abnormal mental health statuses. In addition to traditional hyperparameter tuning techniques, recent studies have explored novel metaheuristic approaches to optimize Random Forest parameters. For instance, Tan et al. \cite{tan2024dung} proposed an improved dung beetle optimizer that refines hyperparameter tuning, further enhancing model performance.

For the multi-class classification task, the same hyperparameter grid was used with a slightly reduced scope to streamline the search process. The weighted F1 score guided model selection across all classes, including Normal, Depression, Anxiety, and Personality Disorder. The optimal model achieved balanced performance across multiple categories, leveraging Random Forests' ability to aggregate predictions from diverse decision trees. 

Random Forests’ inherent feature importance metrics provided additional insights into the most influential predictors for mental health classification. This capability enhances interpretability by highlighting covariates that most strongly influence predictions. The Random Forest models were built using the \texttt{RandomForestClassifier} from \texttt{scikit-learn}. Parameter grids for the number of estimators, maximum depth, and other parameters were evaluated with \texttt{GridSearchCV}. Detailed implementation code, including grid search and evaluation procedures, is available on GitHub.

\paragraph{Light Gradient Boosting Machine (LightGBM)}
Light Gradient Boosting Machine (LightGBM) is a gradient-boosting framework optimized for efficiency and scalability, particularly in handling large datasets and high-dimensional data. Gradient Boosting Machines (GBM) work by sequentially building decision trees, where each new tree corrects the errors made by the previous ones, leading to highly accurate predictions. However, traditional GBM frameworks can be computationally intensive, especially for large datasets \cite{friedman2001greedy}. Unlike traditional Gradient Boosting Machines (GBMs), LightGBM employs a leaf-wise tree growth strategy, which enables deeper splits in dense data regions, enhancing performance by focusing complexity where it is most needed. Additional optimizations, such as histogram-based feature binning, reduce memory usage and accelerate training. These enhancements make LightGBM faster and more resource-efficient than standard GBM implementations, without compromising predictive accuracy \cite{ke2017lightgbm}.

Key hyperparameters tuned for LightGBM included the number of boosting iterations (\texttt{n\_estimators}), learning rate, maximum tree depth (\texttt{max\_depth}), number of leaves (\texttt{num\_leaves}), and minimum child samples (\texttt{min\_child\_samples}). To address the class imbalance, the \texttt{class\_weight} parameter was tested with both \texttt{`balanced'} and \texttt{None} options. Grid search was employed to explore all possible combinations of these hyperparameters, and the weighted F1 score was used as the primary metric for selecting the best configuration.

LightGBM was applied to both binary and multi-class mental health classification tasks. For binary classification, the model differentiated between Normal and Abnormal statuses. For multi-class classification, it predicted categories such as Normal, Depression, Anxiety, and Personality Disorder using the \texttt{multiclass} objective. Hyperparameter tuning via grid search ensured balanced performance across all categories, guided by the weighted F1 score. 

The best-performing models demonstrated robust predictive power, evaluated using precision, recall, F1 scores, confusion matrices, and one-vs-rest ROC curves. Additionally, LightGBM’s feature importance metrics provided interpretability by highlighting the most influential linguistic and sentiment-based features. Its combination of high performance, scalability, and interpretability made LightGBM a key component in this project. The LightGBM models were developed using the \texttt{LGBMClassifier} from the \texttt{lightgbm} library. Hyperparameter tuning, including the number of boosting iterations, learning rate, and tree depth, was performed using \texttt{GridSearchCV}.
Detailed implementation code, including grid search procedures, is available on GitHub.

\subsubsection{A Lite version of Bidirectional Encoder Representations from Transformers (ALBERT)}
A Lite version of Bidirectional Encoder Representations from Transformers (BERT), known as ALBERT \cite{lan2020albert}, is a transformer-based model designed to improve efficiency while maintaining performance. While BERT \cite{devlin2019bert} is highly effective for a wide range of natural language processing (NLP) tasks, it is computationally expensive and memory-intensive due to its large number of parameters. ALBERT addresses these limitations by introducing parameter-sharing across layers and factorized embedding parameterization, which significantly reduces the number of parameters without sacrificing model capacity. Additionally, ALBERT employs Sentence Order Prediction (SOP) as an auxiliary task to enhance pretraining, improving its ability to capture sentence-level coherence. These optimizations make ALBERT a lightweight yet powerful alternative to BERT, capable of achieving competitive performance with reduced memory and computational requirements, making it particularly suitable for large-scale text classification tasks like mental health detection.

In this project, ALBERT was employed for both binary and multi-class classification tasks. For binary classification, the model was fine-tuned to differentiate between Normal and Abnormal mental health statuses, while for multi-class classification, it was configured to predict multiple categories, including Normal, Depression, Anxiety, and Personality Disorder. The implementation leveraged the pre-trained \texttt{Albert-base-v2} model, with random hyperparameter tuning conducted over 10 iterations to optimize the learning rate, number of epochs, and dropout rates. The weighted F1 score served as the primary evaluation metric throughout the tuning process.

For both binary and multi-class classification tasks, hyperparameter tuning was conducted to optimize learning rates between $10^{-5}$ and $10^{-4}$, dropout rates between 0.1 and 0.5, and epochs ranging from 3 to 5. For binary classification, the model achieved high validation F1 scores and demonstrated strong generalization on the test set. For multi-class classification, the objective was adjusted to predict seven categories, with weighted cross-entropy loss applied to address class imbalances and ensure adequate representation of minority categories. The final models were evaluated on the test set using metrics such as accuracy, weighted F1 scores, and confusion matrices.

ALBERT’s architecture efficiently captures long-range dependencies in text while retaining the computational advantages of its lightweight design. The use of random hyperparameter tuning further refined its performance, enabling robust classification for both binary and multi-class tasks. The ALBERT models were fine-tuned with the \texttt{Transformers} (\texttt{AlbertTokenizer} and \texttt{AlbertForSequenceClassification}) library from Hugging Face. Hyperparameter tuning was conducted manually through random search over learning rates, dropout rates, and epochs. etailed implementation code, including data preparation, training, and hyperparameter tuning, is available on GitHub.

\subsubsection{Gated Recurrent Units (GRUs)}
Gated Recurrent Units (GRUs) are a type of recurrent neural network (RNN) designed to capture sequential dependencies in data, making them particularly effective for natural language processing (NLP) tasks such as text classification \cite{cho2014learning}. Compared to Long Short-Term Memory networks (LSTMs), GRUs are computationally more efficient due to their simplified architecture, which combines the forget and input gates into a single update gate. This efficiency allows GRUs to model long-range dependencies while reducing the number of trainable parameters.

In this study, GRUs were employed for both binary and multi-class mental health classification tasks. For binary classification, the model was configured to differentiate between Normal and Abnormal mental health statuses. For multi-class classification, it was adapted to predict categories such as Normal, Depression, Anxiety, and Personality Disorder. 

The GRU architecture comprised three key components:
\vspace{-3mm}
\begin{enumerate}
\item \textbf{Embedding Layer}: Converts token indices into dense vector representations of a fixed embedding dimension.
\vspace{-1.5mm}
\item \textbf{GRU Layer}: Processes input sequences and retains contextual information across time steps, utilizing only the final hidden state for classification.
\vspace{-1.5mm}
\item \textbf{Fully Connected Layer}: Maps the hidden state to output logits corresponding to the number of classes.
\end{enumerate}
\vspace{-3mm}
Dropout regularization was applied to prevent overfitting, and a weighted cross-entropy loss function was used to address class imbalances in the dataset.

For both binary and multi-class classification tasks, hyperparameter tuning was conducted using random search across predefined ranges. The parameters optimized included embedding dimensions (150--250), hidden dimensions (256--768), learning rates ($10^{-4}$--$10^{-3}$), and epochs (5--10). The weighted F1 score served as the primary evaluation metric during validation. The best-performing models achieved high F1 scores on validation datasets and demonstrated robust generalization on the test sets.

GRUs excelled at capturing sequential patterns in text, enabling the model to identify linguistic cues associated with mental health conditions. Despite being less interpretable than tree-based models, their lightweight architecture ensured computational efficiency and strong performance in text-based classification tasks. The GRU models were implemented with the \texttt{torch.nn} module in PyTorch. Key layers included \texttt{nn.Embedding}, \texttt{nn.GRU}, and \texttt{nn.Linear}. Optimization was performed using the \texttt{torch.optim.Adam} optimizer, and class weights were applied using \texttt{nn.CrossEntropyLoss}.
Detailed implementation code, including data preprocessing, model training, and evaluation, is available on GitHub.

\subsection{Evaluation Metrics}
When modeling mental health statuses—particularly for conditions like depression or suicidal ideation—class distributions are often skewed. In many real-world scenarios, the “positive” class (e.g., individuals experiencing depression) is underrepresented compared to the “negative” class (e.g., no mental health issue). This imbalance renders certain evaluation metrics, such as accuracy, less informative: a model that predicts “no issue” for every instance might still achieve high accuracy if the majority class dominates. Consequently, more nuanced metrics are preferred to evaluate the performance of classification models:

\subsubsection{Precision} 
Precision measures the proportion of positive predictions that are truly positive:
\begin{equation}
 \text{Precision} = \frac{\text{True Positives}}{\text{True Positives} + \text{False Positives}}.
\end{equation}
For instance, in depression detection, high precision indicates that most users flagged as “depressed” indeed exhibit depressive content. While precision minimizes false alarms, focusing on it exclusively can be risky. A model that generates very few positive predictions may achieve artificially high precision while missing many genuinely positive cases.

\subsubsection{Recall (Sensitivity)} 
Recall captures the proportion of actual positives correctly identified:
\begin{equation}\nonumber
 \text{Recall} = \frac{\text{True Positives}}{\text{True Positives} + \text{False Negatives}}.
\end{equation}
In depression detection, recall is critical because failing to recognize at-risk individuals (false negatives) can have serious consequences. A model with low recall risks overlooking individuals who need intervention.

\subsubsection{F1 Score} 
The F1 score serves as the harmonic mean of precision and recall, providing a balance between these two metrics \cite{powers2011evaluation}:
\begin{equation}\nonumber
 F1 = 2 \cdot \frac{\text{Precision} \cdot \text{Recall}}{\text{Precision} + \text{Recall}}.
\end{equation}
The F1 score is particularly useful in imbalanced classification scenarios because it penalizes extreme trade-offs, such as very high precision coupled with very low recall. In mental health detection, achieving a high F1 score ensures the model can effectively identify positive cases while maintaining a reasonable level of precision in its predictions.

\subsubsection{Area Under the Receiver Operating Characteristic Curve (AUROC)} 
AUROC provides an aggregate measure of performance across all possible classification thresholds. It evaluates the model's ability to discriminate between positive and negative classes. However, in the presence of severe class imbalance, AUROC may not fully reflect the challenges posed by a majority class dominating the dataset. Nevertheless, it remains valuable for assessing model performance across varying decision thresholds \cite{davis2006relationship}.

\section{Results}
This section presents the findings from the analysis of the dataset and the evaluation of machine learning and deep learning models for mental health classification. First, we provide an \textit{Overview of Mental Health Distribution}, highlighting the inherent class imbalances within the dataset and their implications for model development. Next, the \textit{Hyperparameter Optimization} subsection details the parameter tuning process, which ensures that each model performs at its best configuration for both binary and multi-class classification tasks. Finally, the \textit{Model Performance Evaluation} subsection compares the models' performance based on key metrics, including F1 scores and Area Under the Receiver Operating Characteristic Curve (AUROC). Additionally, nuanced observations, such as the challenges associated with underrepresented classes, are discussed to provide deeper insights into the modeling outcomes.

\subsection{Overview of Mental Health Distribution}
Before hyperparameter optimization and model evaluation, an analysis of the dataset’s class distributions was conducted to highlight potential challenges in classification. The dataset, sourced from Kaggle, contains a total of 51,074 unique statements categorized into three primary groups: \textit{Normal} (31\%), \textit{Depression} (29\%), and \textit{Other} (40\%). The \textit{Other} category encompasses a range of mental health statuses such as \textit{Anxiety}, \textit{Stress}, and \textit{Personality Disorder}, among others.

\begin{figure}[h!]
 \centering
 \includegraphics[width=0.8\textwidth]{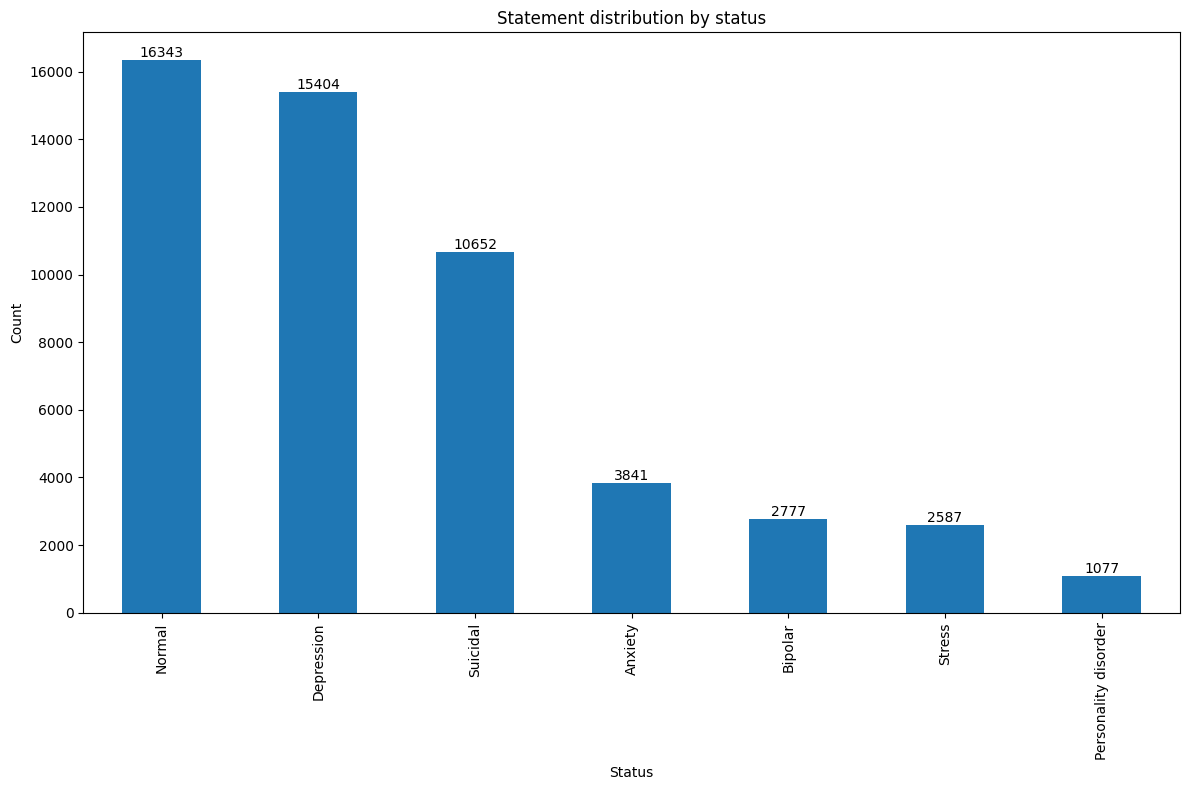}
 \caption{Multi-class distribution of mental health statuses.}
 \label{fig:multi-class}
\end{figure}

\textbf{Figure~\ref{fig:multi-class}} illustrates the expanded distribution of mental health statuses across seven detailed categories in the multi-class classification setup. The dataset shows a significant imbalance, with categories such as \textit{Normal}, \textit{Depression}, and \textit{Suicidal} dominating the distribution, while others like \textit{Stress} and \textit{Personality Disorder} are notably underrepresented. This class imbalance poses challenges for multi-class classification tasks, particularly for the accurate identification of minority classes. Addressing such imbalances requires techniques like class-weighted loss functions and the use of metrics such as weighted F1 scores for model evaluation.

\figurehere{1}

\figurehere{2}

\begin{figure}[h!]
 \centering
 \includegraphics[width=0.8\textwidth]{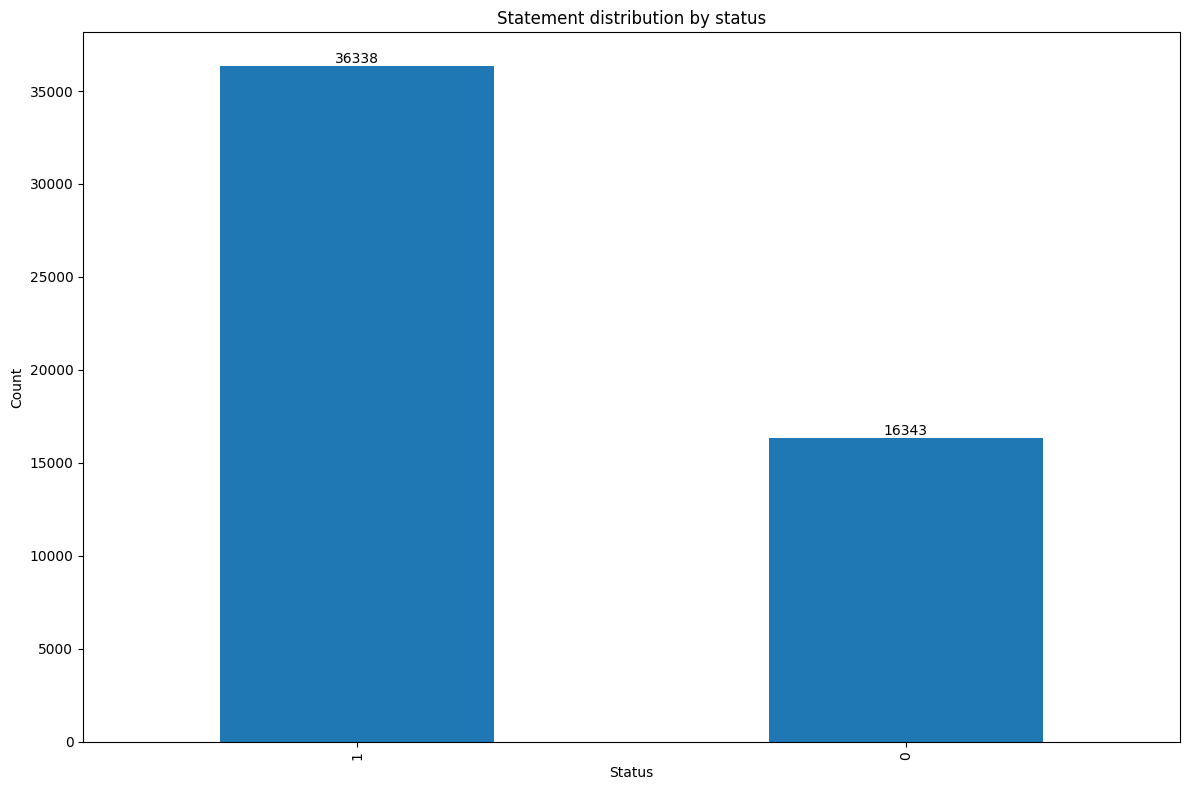}
 \caption{Binary classification distribution of \textit{Normal} versus \textit{Abnormal} mental health statuses.}
 \label{fig:binary-class}
\end{figure}

For the binary classification task, the dataset is divided into two classes: \textit{Normal} and \textit{Abnormal}. The distribution, shown in \textbf{Figure~\ref{fig:binary-class}}, reveals that the \textit{Abnormal} class (labeled as 1) accounts for approximately twice the number of records as the \textit{Normal} class (labeled as 0). Although the imbalance is less severe compared to the multi-class scenario, it still necessitates strategies to ensure that the minority class (\textit{Normal}) is adequately captured during model training.

\subsection{Hyperparameter Optimization}
Hyperparameter optimization is a critical step in enhancing the performance of machine learning (ML) and deep learning (DL) models. For this study, a grid search or random search approach was employed to systematically explore a predefined range of hyperparameters for each model. The primary evaluation metric used to select the best-performing hyperparameter configuration was the weighted F1 score, as it effectively balances precision and recall, particularly in the presence of imbalanced class distributions. This approach ensures that the selected models perform robustly across both binary and multi-class mental health classification tasks. 

The optimized hyperparameters for each model, alongside their corresponding weighted F1 scores on the test set, are summarized in Table~\ref{tbl1:opt_hp}. These results highlight the configurations that achieved the best trade-off between underfitting and overfitting, providing insight into the hyperparameter values critical to the classification tasks.

\tablehere{1}

\subsection{Model Performance Evaluation}

The evaluation metrics, including F1 scores (\textbf{Table~\ref{tbl2:f1-scores}}) and Area Under the Receiver Operating Characteristic Curve (AUROC) (\textbf{Table~\ref{tbl3:auc_scores}}), reveal minimal numeric differences across the models for both binary and multi-class classification tasks. This consistency in performance can be attributed to two primary factors. First, each model underwent rigorous hyperparameter tuning, ensuring only the best configurations were used for evaluation. Second, the dataset size, being of medium volume, provided sufficient information for machine learning models to achieve strong performance, while deep learning models could not fully demonstrate their potential advantages due to the limited data scale.

\tablehere{2}

\tablehere{3}

In the binary classification task, all models exhibited competitive F1 scores and AUROC values, effectively balancing precision and recall while distinguishing between normal and abnormal mental health statuses. Deep learning models such as \textit{ALBERT} and \textit{GRU} demonstrated slightly superior performance, achieving AUROC values of 0.95 and 0.94, respectively, which highlights their ability to capture complex linguistic patterns. Machine learning models, including \textit{Logistic Regression} and \textit{LightGBM}, also performed well, with AUROC scores of 0.93, underscoring their robustness in simpler classification settings.

In the multi-class classification task, a slight decline in performance was observed compared to the binary task. This decline aligns with the increased complexity of distinguishing between seven mental health categories. Nevertheless, deep learning models retained their advantage, with \textit{GRU} and \textit{LightGBM} achieving the highest micro-average AUROC scores of 0.97, followed closely by \textit{ALBERT} with an AUROC of 0.95. Machine learning models such as \textit{Logistic Regression} and \textit{Random Forest} also performed commendably, with AUROC scores of 0.96, demonstrating their ability to handle multi-class tasks effectively when optimized.

Another important observation in the multi-class classification task is the consistently lower AUROC scores for Depression (Class 2) across all machine learning models, with values not exceeding 0.90. While deep learning models demonstrated a slight improvement, their performance for this class remained comparatively weaker than for other categories. This difficulty likely arises from the significant overlap between Depression (Class 2) and other categories in both linguistic and contextual features. The reduced AUROC scores highlight the models' challenges in effectively distinguishing Depression, resulting in higher misclassification rates. These findings emphasize the need for refined feature engineering techniques or more sophisticated model architectures to enhance the separability and accurate classification of this particular class.

The minimal differences in performance metrics across models suggest that the combined effects of comprehensive hyperparameter optimization and dataset size contributed significantly to these results. Binary classification consistently outperformed multi-class classification, likely due to its reduced complexity and fewer decision boundaries. While deep learning models demonstrated their ability to capture intricate patterns, machine learning models offered competitive performance, making them practical alternatives for medium-sized datasets.

Performance metrics for F1 scores and AUROC values are detailed in \textbf{Table~\ref{tbl2:f1-scores}} and \textbf{Table~\ref{tbl3:auc_scores}}, respectively. This analysis highlights the importance of balancing model complexity with dataset characteristics and emphasizes the critical role of hyperparameter tuning in achieving optimal results.

\section{Discussion}

This tutorial serves as a practical resource to address key methodological and analytical challenges in mental health detection on social media, as identified in the systematic review \cite{cao2024mental}. By focusing on best practices and reproducible methods, the tutorial aims to advance research quality and promote equitable outcomes in this important field.

A critical issue identified in the review is the narrow scope of datasets, which are often limited to specific social media platforms, languages, or geographic regions. This lack of diversity restricts the generalizability of findings. In this tutorial, strategies for expanding data diversity are explored, including integrating datasets across multiple platforms, collecting data from underrepresented regions, and analyzing multilingual content. These efforts aim to make research outcomes more inclusive and applicable to diverse populations.

Text preprocessing emerged as another key challenge, particularly in handling linguistic complexities such as negations and sarcasm. These nuances are critical for accurately interpreting mental health expressions. This tutorial offers practical guidelines for building preprocessing pipelines that address these complexities. Techniques for advanced tokenization, feature extraction, and managing contextual meanings are discussed to enhance the reliability of text-based analyses.

Research practices related to model optimization and evaluation were also found to be inconsistent in many studies. Hyperparameter tuning and robust data partitioning are essential for reliable outcomes, yet they are often inadequately implemented. This tutorial provides step-by-step instructions for optimizing models and ensuring fair evaluations, emphasizing the importance of strategies like cross-validation and train-validation-test splits. By following these practices, researchers can reduce bias and improve the validity of their results. 

Evaluation metrics were another area of concern, with many studies relying on accuracy despite its limitations in imbalanced datasets. This tutorial highlights the importance of metrics such as precision, recall, F1-score, and AUROC, which provide a more balanced assessment of model performance. Additionally, practical approaches to managing imbalanced datasets, including oversampling, undersampling, and synthetic data generation, are illustrated.

Transparency in reporting methodologies and results is a foundational element of reproducible research. This tutorial encourages researchers to document their processes comprehensively, including data collection, preprocessing, model development, and evaluation. Sharing code and datasets is also emphasized, fostering collaboration and allowing other researchers to validate findings.

Ethical considerations are central to mental health research, particularly when using sensitive social media data. This tutorial stresses the need for privacy protection and adherence to ethical standards, ensuring that research respects the rights and dignity of individuals. Responsible data handling and clear communication of ethical practices are essential for maintaining trust and accountability in this field.

By addressing these challenges, this tutorial equips researchers with the tools and practices needed to improve the quality and impact of their work. Ultimately, these advancements contribute to the broader goal of promoting equitable and effective mental health interventions on a global scale.

\printbibliography

\newpage
\begin{sidewaystable}[ht]
\centering
\caption{Best Hyperparameters for Binary and Multi-Class Classification Models}
\label{tbl1:opt_hp}
\begin{tabular}{|l|p{5cm}|p{5cm}|p{9cm}|}
\hline
\textbf{Model} & \textbf{Best Parameters (Binary)} & \textbf{Best Parameters (Multi-Class)} & \textbf{Interpretation} \\
\hline
\textbf{Logistic Regression} & 
\texttt{\{C: 10, solver: `liblinear', penalty: `l2', class\_weight: None\}} & 
\texttt{\{C: 10, solver: `lbfgs', penalty: `l2', multi\_class: `multinomial', class\_weight: `balanced'\}} & 
For binary tasks, \texttt{liblinear} is chosen for smaller datasets. For multi-class, \texttt{lbfgs} supports \texttt{`multinomial'} strategy to optimize across multiple categories. Regularization strength (\texttt{C}) of 10 prevents overfitting. \\
\hline
\textbf{SVM} & 
\texttt{\{C: 1, kernel: `rbf', class\_weight: `balanced', gamma: `scale'\}} & 
\texttt{\{C: 1, kernel: `rbf', class\_weight: `balanced', gamma: `scale'\}} &
The RBF kernel captures nonlinear relationships in text data, while \texttt{class\_weight: `balanced'} was selected to address class imbalance. Regularization strength (\texttt{C}) balances margin maximization and misclassification. \\
\hline
\textbf{Random Forest} & 
\texttt{\{n\_estimators: 100, max\_depth: None, min\_samples\_split: 5, min\_samples\_leaf: 1, class\_weight: `balanced'\}} & 
\texttt{\{n\_estimators: 200, max\_depth: None, min\_samples\_split: 2, min\_samples\_leaf: 2, class\_weight: `balanced'\}} & 
For binary tasks, 100 trees ensure stability. For multi-class, 200 trees improve coverage of complex class distributions. Weighted class adjustments handle imbalances. \\
\hline
\textbf{LightGBM} & 
\texttt{\{n\_estimators: 100, learning\_rate: 0.1, max\_depth: -1, num\_leaves: 50, min\_child\_samples: 10, class\_weight: None\}} & 
\texttt{\{n\_estimators: 100, learning\_rate: 0.1, max\_depth: None, num\_leaves: 63, class\_weight: `balanced'\}} & 
For both tasks, LightGBM achieves efficiency via leaf-wise tree growth. For multi-class, additional leaves (63) improve representation of minority classes. \\
\hline
\textbf{ALBERT} & 
\texttt{\{lr: 1.46e-05, epochs: 4, dropout: 0.11\}} & 
\texttt{\{lr: 1.17e-05, epochs: 4, dropout: 0.15\}} & 
ALBERT’s lightweight architecture fine-tunes well with minimal learning rates and dropout for regularization. Minor adjustments improve class representation in multi-class settings. \\
\hline
\textbf{GRU} & 
\texttt{\{embedding\_dim: 156, hidden\_dim: 467, lr: 0.0004, epochs: 5\}} & 
\texttt{\{embedding\_dim: 236, hidden\_dim: 730, lr: 0.0003, epochs: 6\}} & 
Embedding dimensions and hidden states effectively capture sequential dependencies in text. Multi-class configurations benefit from higher hidden dimensions and epochs. \\
\hline
\end{tabular}
\end{sidewaystable}

\begin{table}[htbp]
\centering
\caption{Weighted F1 Scores of Models for Binary and Multi-Class Classification Tasks}
\label{tbl2:f1-scores}
\begin{tabular}{lcc}
\hline
\textbf{Model} & \textbf{Binary Classification} & \textbf{Multi-Class Classification } \\
\hline
Support Vector Machine (SVM) & 0.9401 & 0.7610 \\
Logistic Regression & 0.9345 & 0.7498 \\
Random Forest & 0.9359 & 0.7478 \\
LightGBM & 0.9358 & 0.7747 \\
ALBERT& 0.9576 & 0.7841 \\
Gated Recurrent Units (GRU) & 0.9512 & 0.7756 \\
\hline
\end{tabular}
\end{table}

\begin{table}[ht]
\centering
\caption{Area Under the Receiver Operating Characteristic Curve (AUROC) Scores for Binary and Multi-Class Classification Tasks}
\label{tbl3:auc_scores}
\resizebox{\textwidth}{!}{%
\begin{tabular}{lcc}
\toprule
\textbf{Model} & \textbf{Binary Classification AUROC} & \textbf{Multi-Class Classification Micro-Average AUROC} \\
\midrule
SVM & 0.93 & 0.95 \\
Logistic Regression & 0.93 & 0.96 \\
Random Forest & 0.92 & 0.96 \\
LightGBM & 0.93 & 0.97 \\
ALBERT & 0.95 & 0.97 \\
GRU & 0.94 & 0.97 \\
\bottomrule
\end{tabular}%
}
\end{table}

\newpage

\end{document}